\documentclass[runningheads]{llncs}
\usepackage{booktabs}
\usepackage{multirow}
\usepackage[colorlinks]{hyperref}
\usepackage{amsmath}
\usepackage{graphicx}
\usepackage[misc]{ifsym}

\begin{document}
\title{Dilated Convolutions in Neural Networks for Left Atrial Segmentation in 3D Gadolinium Enhanced-MRI   
}
%
%
\author{Sulaiman Vesal\inst{1}(\Letter)\and
 Nishant Ravikumar\inst{1}\and
 Andreas Maier\inst{1}}
\authorrunning{S. Vesal et al.}
\institute{Pattern Recognition Lab, Friedrich-Alexander-Universit\"at Erlangen-N\"urnberg, Germany\\
\email{sulaiman.vesal@fau.de}}

\maketitle              
\begin{abstract}
Segmentation of the left atrial chamber and assessing its morphology, are essential for improving our understanding of atrial fibrillation, the most common type of cardiac arrhythmia. Automation of this process in 3D gadolinium enhanced-MRI (GE-MRI) data is desirable, as manual delineation is time-consuming, challenging and observer-dependent. Recently, deep convolutional neural networks (CNNs) have gained tremendous traction and achieved state-of-the-art results in medical image segmentation. However, it is difficult to incorporate local and global information without using contracting (pooling) layers, which in turn reduces segmentation accuracy for smaller structures. In this paper, we propose a 3D CNN for volumetric segmentation of the left atrial chamber in LGE-MRI. Our network is based on the well known U-Net architecture. We employ a 3D fully convolutional network, with dilated convolutions in the lowest level of the network, and residual connections between encoder blocks to incorporate local and global knowledge. The results show that including global context through the use of dilated convolutions, helps in domain adaptation, and the overall segmentation accuracy is improved in comparison to a 3D U-Net.

\end{abstract}

\section{Introduction}
Atrial fibrillation (AF) is the most common type of cardiac arrhythmia, with higher rates of incidence among aging populations \cite{Yang}. AF is caused by impaired electrical activity within the atria, which causes cardiac muscle fibers to contract rapidly, in an irregular fashion. The poor performance of current AF treatment strategies results from a lack of adequate understanding of the electrical and structural remodeling characteristics of the human atria \cite{McGann23}\cite{Zhaoe005922}. AF is associated with structural remodeling in the form of fibrosis/scars within the LA and an associated reduction in myocardial voltage. The degree of reduction in voltage and the extent of scar tissue, are indicative of the severity of the pathology \cite{oakes09}.

Current clinical protocol for assessing the left atrium (LA) is electro-anatomi-cal mapping, performed during an electro physiological study. However, this is an invasive technique which uses ionizing radiation, with sub-optimal accuracy, resulting in large localization errors for LA tissue. Gadolinium-enhanced magnetic resonance imaging (GE-MRI) has been shown to improve the visibility of a patient's internal structures, such as the left atria. Consequently, GE-MRI has been widely adopted in recent years, to assess the extent of scar tissue/fibrosis within the LA wall, as a result of AF. While histological analyses have confirmed scar tissue within the LA to be localized to regions of low voltage, quantification of the extent and precise location of the same in a clinical setting, requires invasive techniques. Consequently, the effect of such structural remodeling of cardiac tissue on patient outcome following treatment, and the risk of AF recurrence, are yet to be well understood. \cite{oakes09}. Accurate segmentation of the LA is thus of considerable clinical interest, for diagnosis, treatment planning and patient prognosis. However, manual segmentation of the LA in 3D for structural analysis, is very time consuming, susceptible to manual errors, and is subject to inter-rater differences. Consequently, an automatic approach to LA segmentation is imperative for improved diagnosis and clinical decision making.


Segmentation of the left atrium (LA) in GE-MRI images is a challenging task. Firstly, the LA is hard to distinguish even for expert cardiologists specialized in cardiac MRI. Secondly, respiratory motion, heart rate variability, low signal-to-noise ratio (SNR), and contrast agent wash-out during the long acquisition times, frequently result in poor image quality. \cite{Qian} addressed this challenge using two segmentation steps, namely; an initial global segmentation step using multi-atlas registration; and a subsequent local refinement step based on 3D level-set propagation. However, more recently, convolutional neural networks (CNNs)  have demonstrated state-of-the-art performance in various medical image segmentation tasks. \cite{oolf}  introduced the U-Net architecture, which is trained end-to-end to process the entire image and perform a pixel-wise classification. The V-Net \cite{VNet} architecture, a 3D extension to the U-net, was introduced for volumetric images and enables 3D segmentation, as opposed to processing volumes slice-by-slice. A multi-view CNN was proposed in \cite{Mortazi}, with an adaptive fusion strategy to segment the LA and proximal pulmonary veins in MRI. While, in \cite{8260863} the LA was segmented in computed tomographic angiography (CTA) data, using fully convolutional neural networks (FCNs) with 3D conditional random fields (CRFs), achieving state-of-the-art segmentation accuracy. Other studies \cite{kurzendorfer17} have also investigated automatic segmentation of the left ventricle, in GE-MRI, to assess the extent of scarring resulting from myocardial infarction.

In this paper, we propose a fully automatic 3D segmentation approach for the LA, in AF patients scanned using GE-MRI. Accurate organ segmentation requires incorporation of both local and global information. For this purpose, we constructed a modified version of the 3D U-Net, using dilated convolutions \cite{Dilated} in the lowest layer of the encoder-branch, to extract features spanning a wider spatial range. Additionally, we added residual connections between convolution layers in the encoder branch of the network, to better incorporate multi-scale image information and ensure smoother flow of gradients in the backward pass. A schematic of the proposed network is presented in Fig. \ref{fig.1}. The main contribution of this study is a modified 3D U-Net based segmentation approach (henceforth referred to as 3D U-Net+DR), that incorporates global context via dilated convolutions in the deepest part of the network. We evaluated the LA chamber segmentation accuracy of our network using a data set provided as part of the STACOM MICCAI 2018 challenge, and compared it with a 3D U-Net.


\section{Methodology}
\label{sec:methods}
Incorporating both local and global information is beneficial for many segmentation tasks. In a conventional U-Net however, the lowest level of the network has a small receptive field which prevents the network from extracting features that capture non-local information. Hence, the network would have no understanding that there is only one left atrial cavity within the image, leading to mis-classifications of areas with similar appearance. Dilated convolutions \cite{Dilated} provide a suitable solution to this problem. They introduce an additional parameter, called the dilation rate, to convolution layers, which defines the spacing between values in a kernel. This helps dilate the kernel such that a 3$\times$3$\times$3 kernel with a dilation rate of $2$ will have a receptive field size equal to that of a 7$\times$7$\times$7 kernel. Additionally, this is achieved without any increase in complexity, as the number of parameters associated with the kernel remains the same.
\begin{figure}[!ht]
\label{fig.1}
\centering
\includegraphics[width=\textwidth]{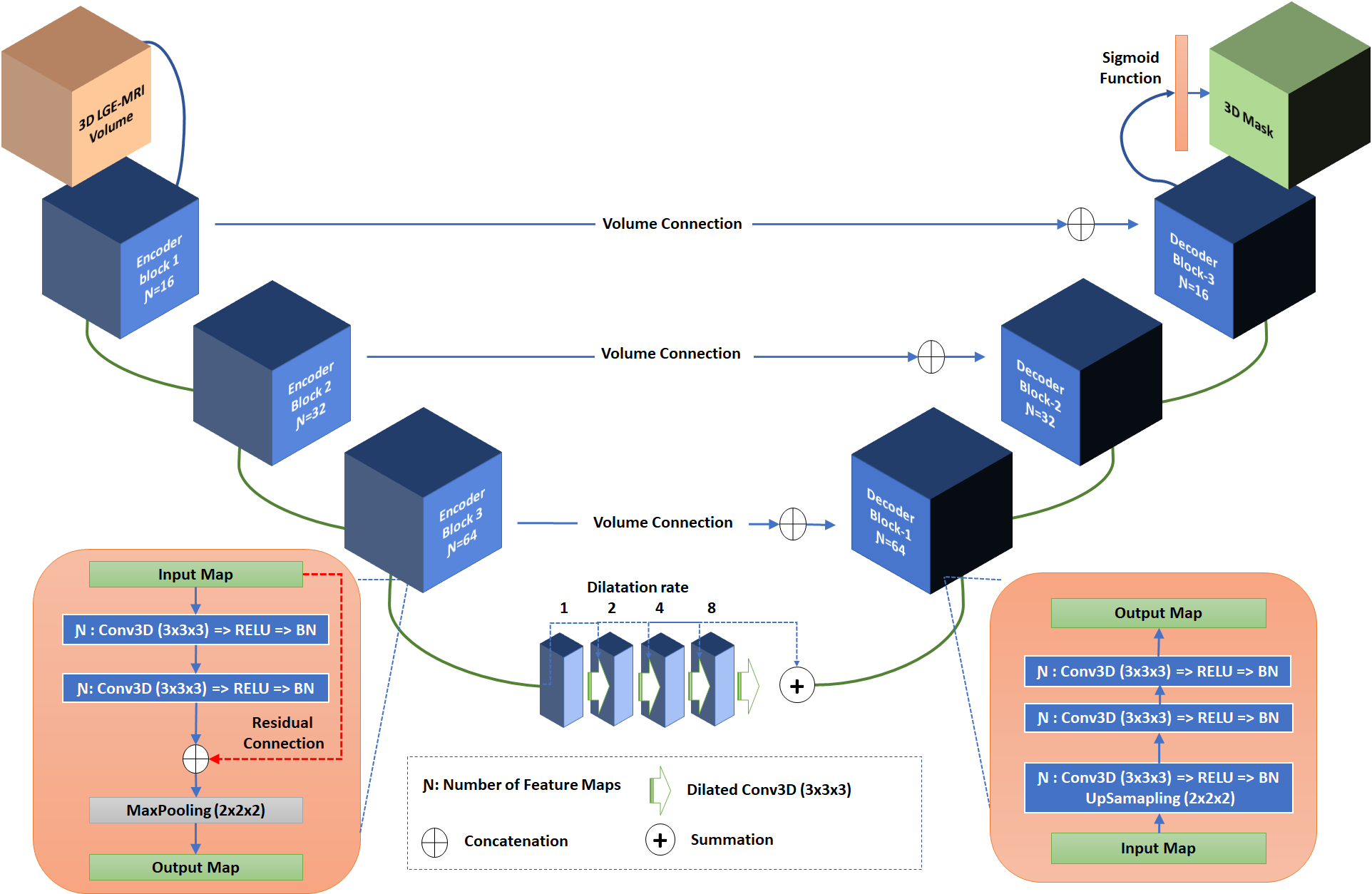}
\caption{3D U-Net+DR network comprising residual connections in the encoder branch and a summation of 4 dilated convolutions layers (with various dilation rates) in the bottleneck.}\label{fig.1}
\end{figure}

The proposed 3D U-Net+DR network (refer to Fig. \ref{fig.1}) comprises three downsampling and upsampling convolution blocks within the encoder and decoder branches, respectively. We use two convolutions with a kernel size of 3$\times$3$\times$3 voxels per block, with batch normalization, rectified linear units (ReLUs) as activation functions, and a subsequent 3D-max pooling operation, as illustrated in Fig. \ref{fig.1}. Image dimensions are preserved between the encoder-decoder branches following convolutions, by zero-padding the estimated feature maps. This enabled corresponding feature maps to be concatenated between the branches. 
A sigomid activation function was used in the last layer to produce a value between $0$ and $1$, to distinguish the background from the foreground. Furthermore, to improve the flow of gradients in the backward pass of the network, the conventional convolution layers in the encoder branch were replaced with residual convolution layers. In each encoder-convolution block, the input to the first convolution layer is concatenated with the output of second convolution layer (red line in Fig. \ref{fig.1}), and the subsequent 3D max-pooling layer reduces volume dimensions by half. The bottleneck between the branches employs four dilated convolutions, with dilation rates $1-4$. The outputs of each were subsequently summed up and provided as input to the decoder branch.

\textbf{Data acquisition:} A total of 100 3D GE-MRIs from patients with AF were provided as part of the STACOM 2018 challenge, for atrial segmentation\footnote{http://atriaseg2018.cardiacatlas.org/}. The resolution of the provided data is $0.625 \times 0.625 \times 0.625 \: mm^3$, with dimensions of $88\times640\times640$ and $88\times576\times576$ voxels. 
Each 3D GE-MRI volume was acquired using a clinical whole-body MRI scanner and its corresponding ground truth binary mask for the LA cavity, was annotated by experts. We split the data set such that 80 volumes were used for training and validating the network via five-fold cross validation, and 20 volumes were used for testing. 

\begin{figure}[tb]
\centering
\includegraphics[width=\textwidth]{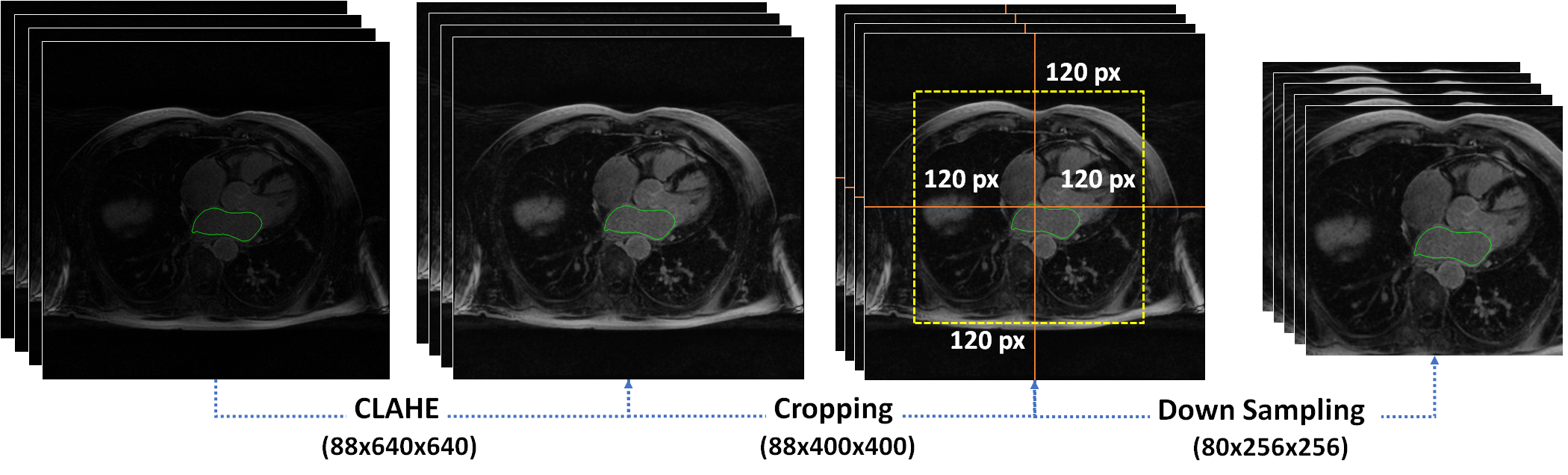}
\caption{Pre-processing workflow: (1) Contrast in GE-MRI volumes is enhanced using CLAHE; (2) resulting volumes are cropped to retain just the ROI; and eventually the cropped volume is down sampled.} \label{fig2}
\end{figure}

\textbf{Data pre-processing:} Due to low contrast in the GE-MRI volumes, we enhanced the contrast slice-by-slice, using contrast limited adaptive histogram equalization (CLAHE), and normalized each volume. Fig. \ref{fig2} illustrates a sample slice image before and after contrast enhancement. In order to retain just the region of interest (ROI), i.e. the LA and its neighboring structures, as the input to our network, each volume was center cropped to a size of 88$\times$400$\times$400. Additionally we downsampled the cropped volumes to 80$\times$256$\times$256, to fit the memory constraints of the GPU (NVIDIA Titan XP 12GB) used for all computations. 

\textbf{Loss function:} The dice coefficient (DC) loss (Eq.\ref{eq1a}) is a measure of overlap widely used for training segmentation networks \cite{VNet}. We used a combination of binary cross entropy (Eq.\ref{eq1b}) and DC loss functions to evaluate network performance. This combined loss (Eq.\ref{eq1c}) is less sensitive to class imbalance in the data and leverages the advantages of both loss functions. Our experiments demonstrated better segmentation accuracy when using the combined loss, relative to employing either individually.  
\begin{subequations}
\begin{equation}
\label{eq1a}
	\zeta_{dc}(y, \hat{y})  = 1- \sum_{k}\frac{\sum_{n}y_{nk} \hat{y}_{nk}}{\sum_{n}y_{nk} + \sum_{n}\hat{y}_{nk}}
\end{equation}
\begin{equation}
\label{eq1b}
\zeta_{bce}(y, \hat{y}) = -\sum_{k}[\hat{y}_{nk}log(\hat{y}_{nk})+(1-y_{nk})(1-\hat{y}_{nk})] 
\end{equation}
\begin{equation}
\label{eq1c}
\zeta(y, \hat{y}) = \zeta_{dc}(y, \hat{y}) + \zeta_{bce}(y, \hat{y})
\end{equation}
\end{subequations}
In Eq.(\ref{eq1a}-\ref{eq1b}) $\hat{y}_{nk}$ denotes the output of the model, where $n$ represents the pixels and $k$ denotes the classes. The ground truth labels are denoted by $y_{nk}$ and we use the two-class version of the DC loss $\zeta_{dc}(y, \hat{y})$ proposed in \cite{VNet}. 

\section{Results and Discussion}
The challenge data set was split into 80\% for training and validation and 20\% for testing. We further split the former into 64 volumes for training and 16 volumes for validation, within a five-fold cross-validation scheme. Networks trained in all experiments were evaluated using the held out test data (20 volumes), and employed the Adam optimizer with a learning rate of 0.001 and 150 epochs.

Three evaluation metrics were used namely: Dice Similarity Coefficient (DC), Jaccard Index (JI) and Accuracy (AC) for the final evaluation of our network. DC and JI measure the degree of overlap between the estimated segmentations and the ground truth masks, while AC measures the proportion of correctly classified pixels. Jaccard is numerically more sensitive to mismatch when there is reasonably strong overlap than DC and AC. The DC, JI and AC measures, averaged across five-fold experiments are summarized in Table \ref{tab1}, for both 3D U-Net and 3D U-Net+DR. The DC scores achieved by the 3D U-Net for the validation and test sets were 81.8\% and 81.8\%, respectively. Our approach on the other hand, achieved DC scores of 85.8\% and 84.8\% for the validation and test sets, respectively. These results indicate that the proposed network significantly outperforms the 3D U-Net, in terms of DC and JI. Additionally, as summarized in the first column of Table \ref{tab1}, 3D U-Net+DR requires just 2.6 million parameters compared to the 3.3 million parameters required by the 3D U-Net. This is due to the summation of the feature maps in the lowest level of the network.
\begin{figure}[tb]
\centering
\includegraphics[width=\textwidth]{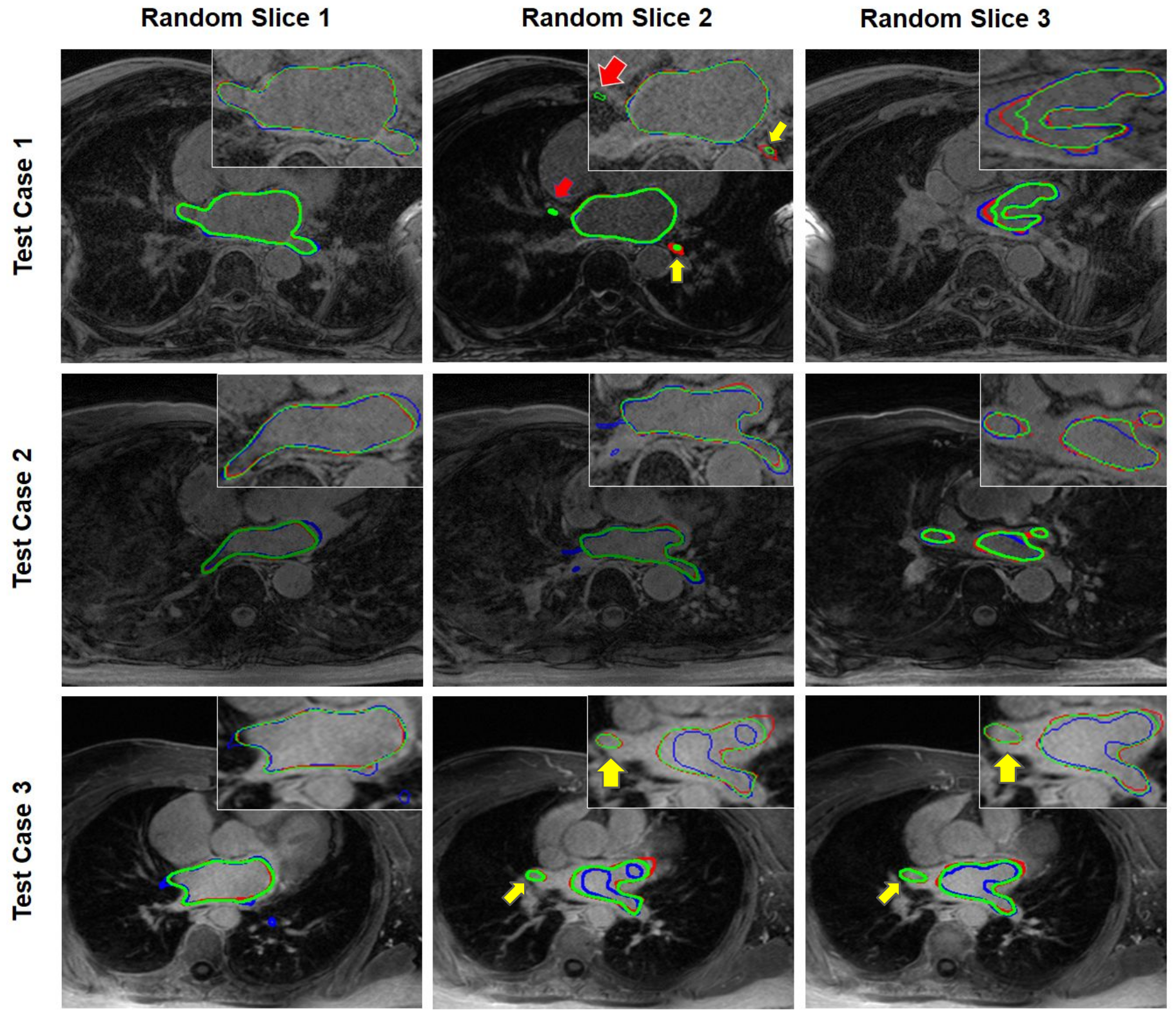}
\caption{Each row shows 3 GE-MRI slices chosen at random from three patients in the test data set. Their corresponding segmentation contours are overlaid: Green represents the ground-truth, while the red and blue contours are estimated by 3D U-Net+DR and 3D U-Net, respectively. The red arrow in the top row, middle slice highlights the true positive tissue areas missed by both methods. The yellow arrows highlight false negatives of 3D U-Net that are captured by 3D U-Net+DR.} \label{fig5}
\end{figure}
\begin{table}[tb]
\label{tab1}
  \centering
  \caption{Segmentation accuracy evaluated in terms of DC, JI and AC metrics, for 3D U-Net and 3D U-Net+DR.}
    \begin{tabular}{p{10.145em}ccccccccc}
    \toprule
    \multicolumn{1}{l}{\multirow{2}[4]{*}{\textbf{Methods}}} & \multicolumn{3}{c}{\textbf{Train Data}} & \multicolumn{3}{c}{\textbf{Validation Data}} & \multicolumn{3}{c}{\textbf{Test Data}} \\
\cmidrule{2-10}    \multicolumn{1}{l}{} & \textbf{AC} & \textbf{DC} & \textbf{JI} & \textbf{AC} & \textbf{DC} & \textbf{JI} & \textbf{AC} & \textbf{DC} & \textbf{JI} \\
    \midrule
    \textbf{3D U-Net\newline{}(3.3 M)} & 0.805     & 0.896     & 0.844     & 0.805     & 0.818     & 0.728     & 0.808     & 0.818     & 0.740 \\
    \midrule
    \textbf{3D U-Net+DR\newline{}(2.6 M)} & \textbf{0.806} & \textbf{0.923} & \textbf{0.873} & \textbf{0.806} & \textbf{0.858} & \textbf{0.768} & \textbf{0.808} & \textbf{0.848} & \textbf{0.770} \\
    \bottomrule
    \end{tabular}%
  \label{tab1}%
\end{table}%

\begin{figure}[tb]
\centering
\includegraphics[width=10cm, height=6.6cm]{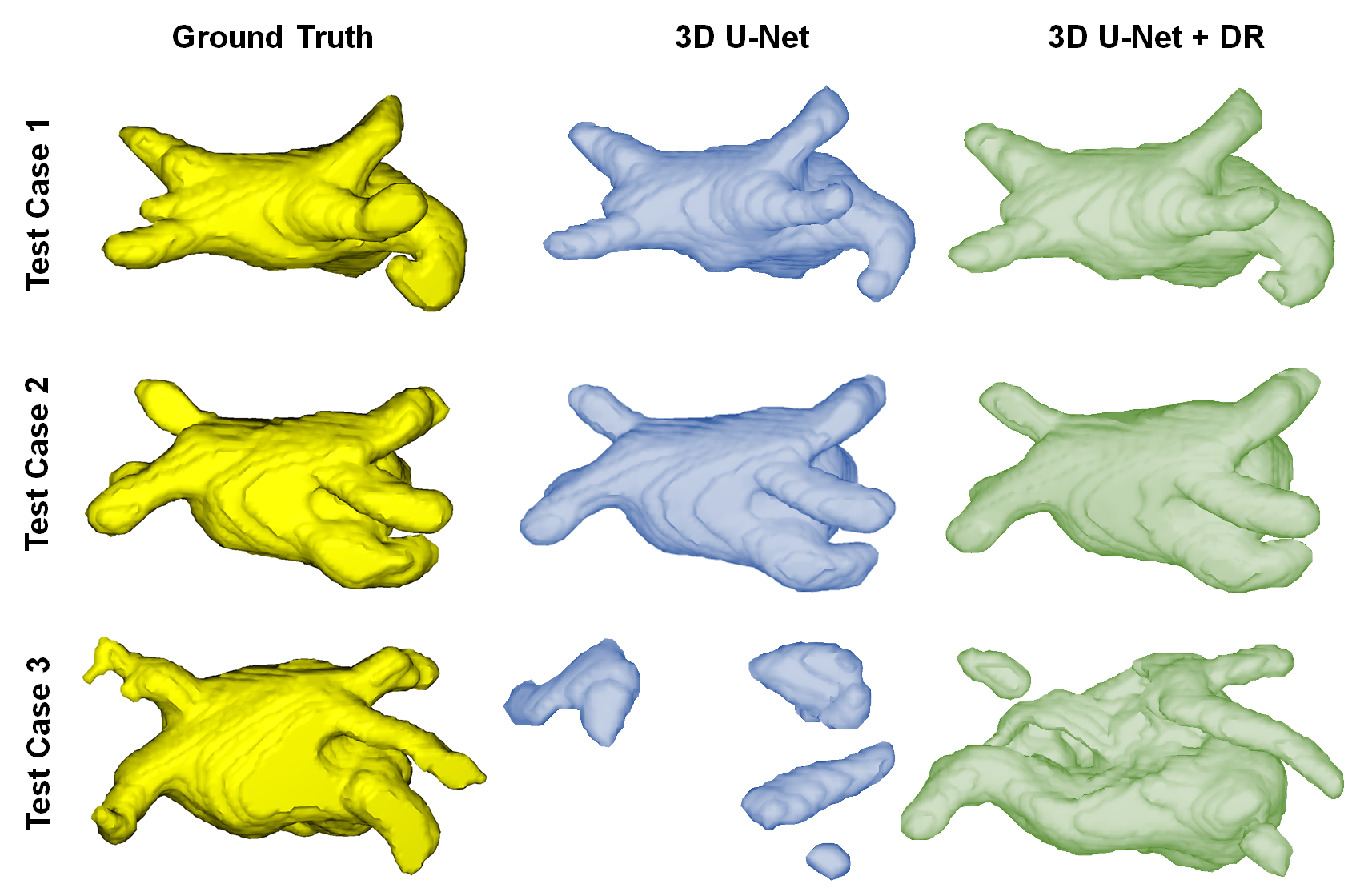}
\caption{3D surface visualization for the ground-truth and the output generated by  3D U-Net and 3D U-Net+DR respectively.} \label{fig6}
\end{figure}

The images presented in Fig. \ref{fig5} help visually assess the segmentation quality of the proposed method on three test volumes. Here, the green contour represents the ground truth segmentation, while the red and blue contours represent the outputs of our method, and 3D U-Net, respectively. These images indicate that the proposed network produces more accurate segmentations, with fewer false negatives than the 3D U-Net. The red arrows in the top row, middle slice highlight atrial regions that both methods missed, while the yellow arrows outline regions where our method succeeded but the 3D U-Net did not. These false negatives of the 3D U-Net were confirmed by visualizing the surface meshes of the estimated segmentations and the ground truth masks for these patient volumes, depicted in Fig. \ref{fig6}. `Test Case 3' presented in the bottom row of Fig. \ref{fig6}, clearly highlights the advantage offered by the use of dilated convolutions in the bottleneck as with our approach, relative to the 3D U-Net. The additional global context imbued within the former, aids in capturing the LA more accurately than the latter.

\section{Conclusion}
\label{sec:conclude}
In this study we proposed a 3D CNN, called 3D U-Net+DR, for automatic segmentation of the LA in GE-MRI volumes of patients diagnosed with AF. Accurate and precise segmentation of the LA is essential for diagnosis, ablation therapy planning and patient prognosis. We leveraged the advantage of training a segmentation network on the complete 3D cardiac MRI volume, rather than 2D slices, to prevent loss of inter-slice information. The proposed method utilizes both local and global information, by expanding the receptive-field in the lowest level of the network, using dilated convolutions. Five-fold cross validation experiments using 100 GE-MRI volumes, revealed significant improvements in segmentation accuracy, evaluated using the DC and JI metrics, were achieved using our approach, compared to a conventional 3D U-Net. In the future we will look to extend our network to incorporate an automatic ROI detection component, within a multi-task learning framework, for improved robustness. 

\bibliographystyle{splncs04}
\bibliography{0000}

\begin{thebibliography}{10}
\providecommand{\url}[1]{\texttt{#1}}
\providecommand{\urlprefix}{URL }
\providecommand{\doi}[1]{https://doi.org/#1}

\bibitem{Yang}
Guang, Y., Xiahai, Z., Habib, K., Shouvik, H., Eva, N., Lei, L., Ricardo, W.,
  Xujiong, Y., Greg, S., Raad, M., Tom, W., Jennifer, K., David, F.: Fully
  automatic segmentation and objective assessment of atrial scars for
  long‐standing persistent atrial fibrillation patients using late
  gadolinium‐enhanced mri. Medical Physics  \textbf{45}(4),  1562--1576
  (2018)

\bibitem{8260863}
Jin, C., Feng, J., Wang, L., Liu, J., Yu, H., Lu, J., Zhou, J.: Left atrial
  appendage segmentation using fully convolutional neural networks and modified
  three-dimensional conditional random fields. IEEE Journal of Biomedical and
  Health Informatics pp.~1--1 (2018)

\bibitem{kurzendorfer17}
Kurzendorfer, T., Forman, C., Schmidt, M., Tillmanns, C., Maier, A., Brost, A.:
  Fully automatic segmentation of left ventricular anatomy in 3-d lge-mri.
  Computerized Medical Imaging and Graphics  \textbf{59},  13--27 (2017)

\bibitem{McGann23}
McGann, C., Akoum, N., Patel, A., Kholmovski, E., Revelo, P., Damal, K.,
  Wilson, B., Cates, J., Harrison, A., Ranjan, R., Burgon, N.S., Greene, T.,
  Kim, D., DiBella, E.V., Parker, D., MacLeod, R.S., Marrouche, N.F.: Atrial
  fibrillation ablation outcome is predicted by left atrial remodeling on mri.
  Circulation: Arrhythmia and Electrophysiology  \textbf{7}(1),  23--30 (2014)

\bibitem{VNet}
Milletari, F., Navab, N., Ahmadi, S.A.: V-net: Fully convolutional neural
  networks for volumetric medical image segmentation. In: 2016 Fourth
  International Conference on 3D Vision (3DV). pp. 565--571 (Oct 2016)

\bibitem{Mortazi}
Mortazi, A., Karim, R., Rhode, K., Burt, J., Bagci, U.: Cardiacnet:
  Segmentation of left atrium and proximal pulmonary veins from mri using
  multi-view cnn. In: Descoteaux, M., Maier-Hein, L., Franz, A., Jannin, P.,
  Collins, D.L., Duchesne, S. (eds.) Medical Image Computing and
  Computer-Assisted Intervention − MICCAI 2017. pp. 377--385. Springer
  International Publishing, Cham (2017)

\bibitem{oakes09}
Oakes, R.S., Badger, T.J., Kholmovski, E.G., Akoum, N., Burgon, N.S., Fish,
  E.N., Blauer, J.J., Rao, S.N., DiBella, E.V., Segerson, N.M., et~al.:
  Detection and quantification of left atrial structural remodeling with
  delayed-enhancement magnetic resonance imaging in patients with atrial
  fibrillation. Circulation  \textbf{119}(13),  1758--1767 (2009)

\bibitem{Qian}
Qian, T., Gucuk, I.E., Rahil, S., F., B.F., Saman, N., van~der Geest Rob~J.:
  Fully automatic segmentation of left atrium and pulmonary veins in late
  gadolinium‐enhanced mri: Towards objective atrial scar assessment. Journal
  of Magnetic Resonance Imaging  \textbf{44}(2),  346--354

\bibitem{oolf}
Ronneberger, O., Fischer, P., Brox, T.: U-net: Convolutional networks for
  biomedical image segmentation. In: Navab, N., Hornegger, J., Wells, W.M.,
  Frangi, A.F. (eds.) Medical Image Computing and Computer-Assisted
  Intervention -- MICCAI 2015. pp. 234--241. Springer International Publishing,
  Cham (2015)

\bibitem{Dilated}
Yu, F., Koltun, V.: Multi-scale context aggregation by dilated convolutions.
  In: ICLR (2015)

\bibitem{Zhaoe005922}
Zhao, J., Hansen, B.J., Wang, Y., Csepe, T.A., Sul, L.V., Tang, A., Yuan, Y.,
  Li, N., Bratasz, A., Powell, K.A., Kilic, A., Mohler, P.J., Janssen, P.M.L.,
  Weiss, R., Simonetti, O.P., Hummel, J.D., Fedorov, V.V.: Three-dimensional
  integrated functional, structural, and computational mapping to define the
  structural {\textquotedblleft}fingerprints{\textquotedblright} of
  heart-specific atrial fibrillation drivers in human heart ex vivo. Journal of
  the American Heart Association  \textbf{6}(8) (2017)

\end{thebibliography}

\end{document}